%% file: main.tex
\documentclass{article}
\usepackage{iclr2027_conference,times}
\input{math_commands.tex}

\usepackage{booktabs}
\usepackage{graphicx}
\usepackage{capt-of}
\usepackage{placeins}
\usepackage{hyperref}
\usepackage{url}

\title{TabletVLM: End-to-End Cuneiform Tablet Transcription with a Spatially Grounded Vision Language Model}

\author{Jerry Cheng$^{*}$\\
University of California, Berkeley\\
\texttt{jerrycheng03@berkeley.edu}
\And Tiantong Lu$^{*}$\\
University of California, Berkeley\\
\texttt{lutt@berkeley.edu}
\AND Adam G. Anderson\\
Research Director, FactGrid Cuneiform Project\\
\texttt{adam.anderson@factgrid.eu}}

\iclrfinalcopy
\begin{document}

\maketitle
\lhead{Preprint}
{\renewcommand{\thefootnote}{\fnsymbol{footnote}}\footnotetext[1]{Equal contribution.}}

\begin{abstract}
Transcription of cuneiform tablets requires solving spatial localization, reading-order reconstruction, and sign recognition across substantial variations in image quality, tablet condition, and scribal style. Most current approaches are detection-based, relying on sign-level bounding-box supervision, but existing cuneiform datasets often contain incomplete box annotations and inconsistent sign labels, limiting the reliable supervision available for training in this already low-resource domain. Obtaining additional sign-level annotations requires costly and time-consuming manual work. End-to-end image-to-sequence models offer an alternative by learning transcription directly from paired images and text; however, learning sign appearance, spatial alignment, and sequence structure jointly from limited full-tablet supervision remains challenging. We propose TabletVLM, a vision-language model for cuneiform transcription, and investigate how various training curricula can help improve convergence and transcription performance. On 4,805 held-out tablet-side images, the full curriculum achieved a pooled sign-error rate of $30.01 \pm{1.03\%}$, compared with $77.08 \pm{0.05\%}$ for the recent eBL detector fine-tuned on our data, and $105.20\%$ for TabletVLM trained without curriculum pretraining. On 879 held-out annotated signs, the full curriculum achieves $70.99 \pm{2.09\%}$ peak-in-box localization during free generation. More broadly, our results suggest that end-to-end VLMs trained with structured curricula can help address limited and uneven supervision in low-resource visual recognition tasks.
\end{abstract}

\section{Introduction}

Cuneiform tablets preserve a large textual record of the ancient Near East, but much of this material remains difficult to access computationally without expert transcription. We focus on texts from the Third Dynasty of Ur (Ur III), conventionally dated to approximately 2112-2004 BC, whose large administrative corpus provides a comparatively well-defined setting for studying automated cuneiform transcription. Improving transcription from tablet images could expand the amount of searchable text available for both philological research and large-scale computational analysis.

\subsection{Challenges}
Transcription of cuneiform tablets is difficult due to the complexity and variance in tablets. Signs are formed by pressing a stylus into wet clay, introducing three-dimensional warping not normally encountered in OCR or HTR. As opposed to type-print, cuneiform sign appearance varies by scribal style, and spacing between signs and lines are highly irregular. Roughly four millennia of aging have also subjected Ur III tablets to erosion, cracking, or discoloration which obscure signs and introduce similar visual noise.

\begin{figure}[!htbp]
\centering
\begin{minipage}[t]{0.48\textwidth}
\vspace{0pt}
\centering
\includegraphics[height=0.8\linewidth]{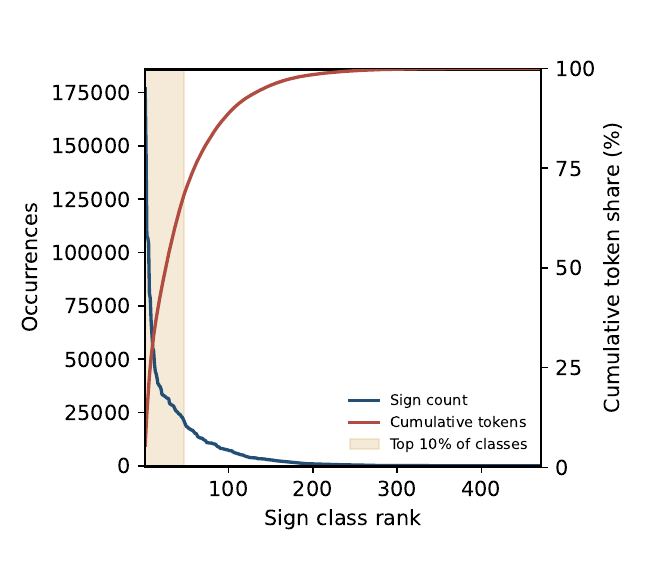}
\captionof{figure}{Sign occurrences by class rank (left axis) and cumulative percentage of tokens (right axis). Shading marks the top 10\% of classes.}
\label{fig:sign-class-distribution}
\end{minipage}\hfill
\begin{minipage}[t]{0.48\textwidth}
\vspace{0pt}
\centering
\includegraphics[height=0.8\linewidth]{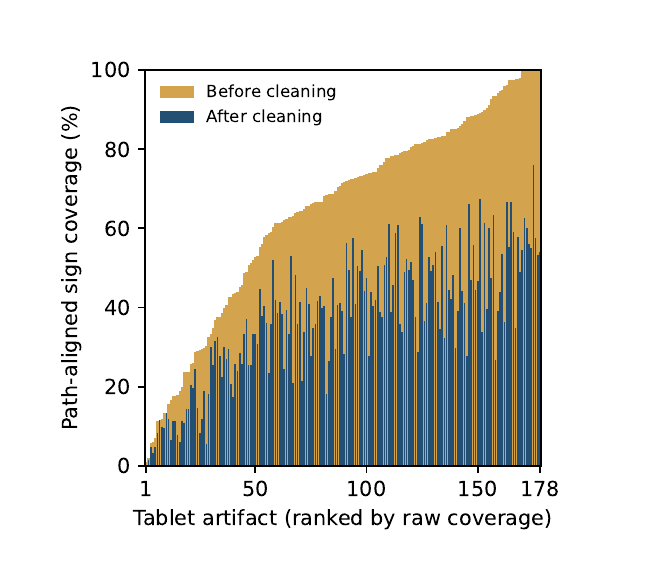}
\captionof{figure}{Path-aligned coverage for 178 artifacts, sorted by raw coverage. Gold and blue show coverage before and after cleaning.}
\label{fig:box-coverage-tablet}
\end{minipage}
\end{figure}

\textbf{Class imbalance} Sign classes also have an extreme long-tailed distribution. In our unified dataset, 471 sign classes appear 3,160,375 times, but the top 10\% of classes (47 of 471) account for 68.28\% of those occurrences, while 67 classes (14.23\%) appear fewer than ten times (Figure~\ref{fig:sign-class-distribution}). This makes it difficult for models to learn rare signs and generalize effectively. Moreover, a sign may occur in tablet transcriptions without an individually annotated image example. Aggregate transcription scores can obscure these weaknesses because frequent signs dominate reference tokens, so we also report macro sign F1, which gives equal weight to each sign class. 

\textbf{Spatial annotations} Spatial annotations are uneven as well. On the eBL tablet sides represented by Sign Snippets~\citep{lewenstein2026cuneiform} retained for this study, raw boxes align with 56.0\% of visible transcribed sign positions, falling to 30.7\%, 4,520 crops spanning 167 sign classes, after the data cleaning and sign mapping (Section~\ref{sec:data-sources}; Figure~\ref{fig:box-coverage-tablet}). These inconsistencies likely reflect the aggregation of manually annotated data from heterogeneous sources with differing annotation conventions and levels of completeness. Expanding or correcting these annotations would require time-consuming manual work by domain experts, making it difficult to obtain reliable sign-level supervision at the scale needed to train robust detection models. 

\textbf{OCR} Current approaches often decompose localization, recognition, and sequencing as separate steps. This increases control and interpretability, but sparse sign boxes leave visible signs unlabeled during detector training \citep{che2026automated}, and errors in localization then propagate to recognition and sequencing. Separately optimized stages also do not naturally share learned representations \citep{williams2023deepscribe}. On the other hand, direct end-to-end training on full tablet images avoids box-level targets \citep{kim2022donut}, but introduces substantial visual and sequential variability. Our experiments show that direct training yields substantially poorer transcription accuracy than training with intermediate supervision, highlighting the complexity of the task. 
\subsection{Our Approach}

We investigate a single vision-language model that jointly learns visual sign recognition and sequence generation to transcribe complete tablet-side images, with multiple training curricula used to evaluate how different sources of supervision support these capabilities. Our model consists of an ImageNet-initialized SwinV2-Tiny backbone, a projection layer, and a three-layer Transformer decoder initialized from scratch, which uses self- and cross-attention to autoregressively generate cuneiform text. 

We use labeled individual sign images, full tablet-side images, and tablet text sequences. Following the synthetic data generation introduced in concurrent work \citep{anonymous2026compositional}, we also construct synthetic training examples by combining visual sign instances with text sequences, providing intermediate supervision between isolated sign recognition and full tablet transcription. We compare curricula that incorporate different combinations of sign-level learning, text-only warm-up, and synthetic line and tablet-side training before real tablet-side transcription. These comparisons examine whether intermediate training stages improve transcription accuracy and spatial alignment over simpler training schedules.

\subsection{Results \& Contributions}
Our model demonstrates that a single vision-language model can learn both transcription and spatial grounding without explicit sign-box targets during full-tablet training. On 4,805 held-out tablet sides, the full curriculum achieves a pooled SER of $30.01 \pm{1.03\%}$ and macro sign F1 of $54.58 \pm{0.83\%}$ across three seeds, compared with $77.08 \pm{0.05\%}$ pooled SER for the fine-tuned eBL detector \citep{che2026automated}. During free generation, attention peaks fall inside the corresponding sign regions for $70.99 \pm{2.09\%}$ of 879 held-out annotated signs. Controlled occlusion provides complementary evidence that predictions depend on the corresponding image regions.

Our contributions are threefold: (1) we develop a unified vision-language model for direct cuneiform tablet-side transcription; (2) we investigate multiple training curricula to assess how different sources and sequences of supervision support learning within this architecture; and (3) we evaluate transcription on 4,805 held-out tablet-side images and spatial alignment through a separate localization evaluation using annotated sign regions.

\section{Related Work}

\textbf{Cuneiform sign detection and localization.}
One line of cuneiform OCR research focuses on locating and identifying individual signs in tablet photographs or renderings \citep{dencker2020cuneiform,stotzner2023cuneiform,cobanoglu2024sign}. These methods provide explicit spatial predictions but require additional processing to recover ordered transcriptions. Incomplete annotations also complicate evaluation, as correct detections of unannotated signs are counted as false positives.

\textbf{Cuneiform sign recognition.}
Isolated recognition methods classify cropped sign images \citep{verwimp2025signs} or align prototypes with photographed signs, using synthetic examples to improve rare-class recognition \citep{mikulinsky2025protosnap}. These methods learn sign identity but do not evaluate localization, reading order, or complete tablet transcription.

\textbf{Cuneiform OCR pipelines.}
Modular pipelines combine explicit detection, recognition, and ordering stages. DeepScribe uses separately trained sign detection, classification, and line detection \citep{williams2023deepscribe}; other approaches combine tablet-side extraction, DETR, and heuristic reading-order reconstruction \citep{che2026automated}, or SAM2 segmentation with a Vision Transformer classifier \citep{poudel2026zeroshot}. These pipelines provide interpretable intermediate outputs, but allow upstream errors to propagate and lack joint optimization for the final transcription objective.

\textbf{End-to-end image-to-sequence OCR.}
Recent OCR models increasingly map document images to ordered text or structured sequences using vision-language models. HunyuanOCR \citep{hunyuan2025ocr} follows a unified image-to-sequence approach, and PaddleOCR-VL \citep{cui2025paddleocrvl} combines a separate layout and reading-order model with a VLM for element recognition. These models typically rely on large quantities of paired image-text supervision. Cuneiform presents a more data-constrained setting in which such supervision is limited and heterogeneous, motivating our investigation of how different sources of supervision support full tablet-side transcription.

\textbf{Synthetic and compositional supervision.}
Synthetic training data provides a way to construct paired image-sequence supervision when manually aligned examples are scarce. StackMix \citep{shonenkov2021stackmix} generates labeled handwritten sequences by recombining smaller visual fragments, demonstrating that supervision available at a local scale can be composed into longer recognition examples. More recently, \citet{li2026realworld} construct realistic full-page document supervision by composing document elements and layouts, using synthetic data to address the scarcity of end-to-end document annotations. Closely related concurrent work~\citep{anonymous2026compositional} applies this principle to cuneiform by arranging isolated real sign crops according to tablet text sequences to produce synthetic line-level supervision. We utilize synthetic line and tablet images to bridge the gap between sign images and tablet images. 

\textbf{Progressive and curriculum-based training.}
Curriculum learning organizes training so that simpler tasks are learned before more difficult ones \citep{bengio2009curriculum}. Recent document-recognition systems have adopted similar strategies: \citet{li2026realworld} combine synthetic full-page supervision with progressive learning for end-to-end document parsing, while FireRed-OCR \citep{wu2026fireredocr} trains a VLM through successive stages that move from document perception toward structured sequence generation. Our curriculum follows the same general principle but increases difficulty along spatial and sequential dimensions, requiring increasingly complex recognition, localization, and reading-order reconstruction while retaining a single model throughout training.
\section{Dataset}

\subsection{Data Sources and Normalization}
\label{sec:data-sources}

We build a unified Ur III dataset with sign crops from Sign Snippets~\citep{lewenstein2026cuneiform}, sign annotations from the electronic Babylonian Library (eBL)~\citep{ebl}, and tablet photographs, catalog metadata, and ASCII Text Format (ATF) transcriptions from the Cuneiform Digital Library Initiative (CDLI)~\citep{cdli2026}. These sources provide sign-level and tablet-level supervision. We normalize their labels to a shared Unicode cuneiform vocabulary.

\paragraph{Data Cleaning \& Filtering} 
We apply strict cleaning and filtering criteria using online catalog definitions and manual inspection. For Sign Snippets, we retain consistent labels that map to Unicode cuneiform, identify source tablet using CDLI IDs, and obtain individual crops from annotated sign regions (Appendix~\ref{app:sign-snippets}). For CDLI, we match photographs and ATF records by catalog identity, parse side and line structure, and automatically extract obverse and reverse images from composite photographs, discarding cases without two eligible regions (Appendix~\ref{app:dataset-construction}).

\paragraph{Sign Annotation Coverage}
Sign annotations cover only part of the visible text in the eBL images. Across 206 artifacts containing 14,610 visible transcribed sign positions, annotation coverage is 56.0\% in the raw metadata and 30.7\% after filtering. Figure~\ref{fig:box-coverage-tablet} shows coverage before and after filtering for the 178 artifacts with nonzero coverage; 28 zero-coverage artifacts are omitted from the plot but included in the overall rates. Coverage is measured by matching boxes to transcription positions, so unmatched positions may also reflect outdated transcription paths (Appendix~\ref{app:sign-snippets}).

\paragraph{Synthetic Data}
Following the synthetic data generation procedure introduced in concurrent work~\citep{anonymous2026compositional}, we construct synthetic line and tablet-side training examples from these independently available resources, bridging isolated sign recognition and full tablet-side transcription (Appendix~\ref{app:synthetic-rendering}).

\paragraph{Target}
The target is an ordered sequence of Unicode signs with line, column, and uncertainty markers. The training inventory contains 471 signs and 8 special tokens. Unicode mapping details are in Appendix~\ref{app:unicode-normalization}. Figure~\ref{fig:sign-class-distribution} shows the sign-class distribution across the unified manifest.

\subsection{Dataset Splits and Provenance}

We use CDLI P-numbers to enforce source-tablet isolation. The transcription test set contains 2,485 tablets, approximately 10\% of the cleaned CDLI tablet pool. A separate localization evaluation holdout contains 67 of 133 eligible eBL tablets with sign annotations. After excluding both sets from the training pool, we assign 10\% of the remaining source P-numbers to validation by deterministic SHA-256 ranking with seed 13. Synthetic examples use donors from their assigned split. No source tablet appears in more than one of the training, validation, or test splits, and validation and test set vocab is fully covered by the train set (Appendix~\ref{app:source-accounting}).

\begin{table}[!htbp]
\centering
\caption{Dataset composition after normalization, source-level splitting, and provenance filtering.}
\label{tab:dataset_statistics}
\begin{tabular}{lrrr}
\toprule
\textbf{Component} &
\textbf{Train} &
\textbf{Validation} &
\textbf{Test} \\
\midrule
Real sign crops
    & 4,520 & 444 & -- \\
Synthetic lines
    & 140,981 & 4,437 & -- \\
Synthetic tablet sides
    & 17,623 & 555 & -- \\
Real CDLI tablet sides
    & 38,920 & 4,256 & 4,805 \\
\midrule
\textbf{Total}
    & \textbf{202,044}
    & \textbf{9,692}
    & \textbf{4,805} \\
\bottomrule
\end{tabular}
\end{table}

\FloatBarrier

\section{Model Design}
Our model follows an encoder-projector-decoder architecture for autoregressive cuneiform transcription. The encoder extracts spatial visual features from the tablet image, and the projector maps these features into the decoder's embedding space. The decoder then generates a sequence of Unicode cuneiform signs and structural markers, conditioning each output token on the image tokens and previously generated tokens. 

\paragraph{Encoder}
The encoder uses SwinV2-Tiny pretrained on ImageNet-1K~\citep{liu2022swinv2}. Its four stages progressively reduce spatial resolution while increasing feature dimensionality, producing a final feature map at $1/32$ of the input height and width. In preliminary experiments, the model was unable to converge during tablet-side training, suggesting that reduced spatial resolution may obscure the detail needed to recognize individual signs.

We therefore disable spatial downsampling in the fourth Swin stage during Phases 3--5. This preserves feature maps at $1/16$ of the input height and width, yielding four times as many spatial positions as the standard configuration while retaining 768 feature channels. Table~\ref{tab:swin-dimensions} compares the feature dimensions, and Appendix~\ref{app:swin-downsampling} explains the modification.

\FloatBarrier
\begin{table}[!htbp]
\centering
\caption{Swin feature dimensions (height $\times$ width $\times$ channels)
with standard downsampling and with Stage 4 spatial downsampling disabled.}
\label{tab:swin-dimensions}
\begin{tabular}{lcc}
\hline
Swin stage & Standard Swin & Stage 4 at /16 \\
\hline
Input           & $512 \times 1024 \times 3$ & $512 \times 1024 \times 3$ \\
Patch embedding & $128 \times 256 \times 96$ & $128 \times 256 \times 96$ \\
Stage 1         & $128 \times 256 \times 96$ & $128 \times 256 \times 96$ \\
Stage 2         & $64 \times 128 \times 192$ & $64 \times 128 \times 192$ \\
Stage 3         & $32 \times 64 \times 384$  & $32 \times 64 \times 384$ \\
Stage 4         & $16 \times 32 \times 768$  & $32 \times 64 \times 768$ \\
\hline
\end{tabular}
\end{table}
\FloatBarrier

\paragraph{Projector \& Decoder}
Separate projectors map 768-channel features through widths \(768\to384\to192\) for lines and \(768\to192\to192\) for sides. The spatial grid is represented as a sequence of visual tokens. With Stage 4 spatial downsampling disabled, the full grid contains 2,048 visual tokens, each with 192 dimensions:

\[
32 \times 64 \times 768
\;\rightarrow\;
32 \times 64 \times 192
\;\rightarrow\;
2048 \times 192.
\]

The decoder is a three-layer autoregressive Transformer with a hidden dimension of 192, three attention heads, and a feed-forward dimension of 768. Each layer contains causal self-attention over previous tokens, cross-attention over the projected image tokens, and a feed-forward network.

A linear output layer maps each decoder hidden state to a score for every vocabulary token, sharing its weights with the token embedding matrix. The vocabulary is constructed from the training transcriptions and includes Unicode cuneiform signs, the structural tokens \texttt{<LINE>}, \texttt{<COLUMN>}, \texttt{<ILLEGIBLE>}, and \texttt{<GAP>}, and model-control tokens. Autoregressive generation begins with \texttt{<BOS>} and stops when the model predicts \texttt{<EOS>} or reaches the 512-token generation limit. This limit corresponds to the 99.17th percentile across the whole CDLI transcription population.

The model contains 30.67 million parameters, of which 29.75 million are used for tablet-side transcription, including 27.58 million in the encoder and 1.97 million in the decoder.

\section{Training}
\label{sec:training-curricula}

We study six training curricula assembled from the same five fixed phase definitions. Each curriculum is an independent run, model weights are carried between phases within a run. Phase-5-only training is retained as a single-run control at seed 13, the rest are evaluated across seeds 13, 42, and 123. The six curricula are:

\begin{center}
\begin{tabular}{lll}
\hline
Curriculum & Phase sequence & Seeds \\
\hline
Full curriculum & \(1 \rightarrow 2 \rightarrow 3 \rightarrow 4 \rightarrow 5\) & 13, 42, 123 \\
Without synthetic transcription & \(1 \rightarrow 2 \rightarrow 5\) & 13, 42, 123 \\
Text warm-up followed by real sides & \(2 \rightarrow 5\) & 13, 42, 123 \\
Without synthetic tablet sides & \(1 \rightarrow 2 \rightarrow 3 \rightarrow 5\) & 13, 42, 123 \\
Without synthetic lines & \(1 \rightarrow 2 \rightarrow 4 \rightarrow 5\) & 13, 42, 123 \\
Real sides only & \(5\) & 13 \\
\hline
\end{tabular}
\end{center}

All curricula use the same dataset manifest and split and vocabulary of 479 tokens. Synthetic-line and synthetic-tablet training sets are each capped at 10,000 randomly selected examples to prevent overfitting, while their validation sets remain uncapped. Each run starts with an ImageNet-pretrained SwinV2-Tiny encoder and randomly initialized decoder and projection layers. Training follows the phase definitions in Table~\ref{tab:training-phases}, except that curricula \(2\to5\) and \(5\) train the patch embedding and all Swin stages during Phase 5.

\begin{table}[tbp]
\centering
\caption{Training phases with representative training examples, trained components, and frozen Swin encoder stages.}
\label{tab:training-phases}
\setlength{\tabcolsep}{3pt}
\begin{tabular}{@{}p{0.065\textwidth}p{0.145\textwidth}p{0.16\textwidth}p{0.20\textwidth}p{0.20\textwidth}p{\dimexpr0.23\textwidth-10\tabcolsep\relax}@{}}
\toprule
\textbf{Phase} & \textbf{Data} & \textbf{Visual} & \textbf{Objective} & \raggedright\textbf{Trained components} & \raggedright\textbf{Frozen stages} \tabularnewline
\midrule
1 & \raggedright Isolated sign crops
  & \vspace{0pt}\includegraphics[width=\linewidth,height=0.55in,keepaspectratio]{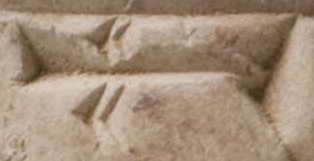}
  & \raggedright Sign classification
  & \raggedright Encoder, classifier
  & \raggedright None \tabularnewline
\midrule
2 & \raggedright Training transcriptions (text only)
  & \vspace{0pt}\includegraphics[width=\linewidth,height=0.55in,keepaspectratio]{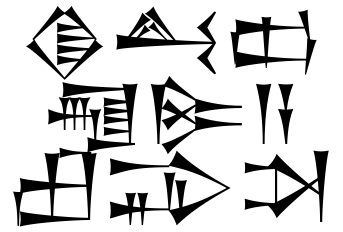}
  & \raggedright Causal next-token prediction
  & \raggedright Decoder (cross-attention bypassed)
  & \raggedright Patch embedding; Stages 1--4 \tabularnewline
\midrule
3 & \raggedright Synthetic lines
  & \vspace{0pt}\includegraphics[width=\linewidth,height=0.55in,keepaspectratio]{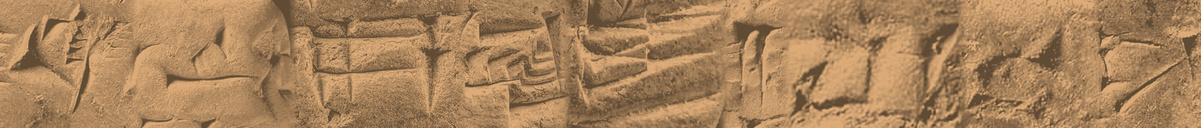}
  & \raggedright Autoregressive line transcription
  & \raggedright Encoder, decoder, line projector
  & \raggedright Patch embedding; Stages 1--3 \tabularnewline
\midrule
4 & \raggedright Synthetic tablet sides
  & \vspace{0pt}\includegraphics[width=\linewidth,height=0.55in,keepaspectratio]{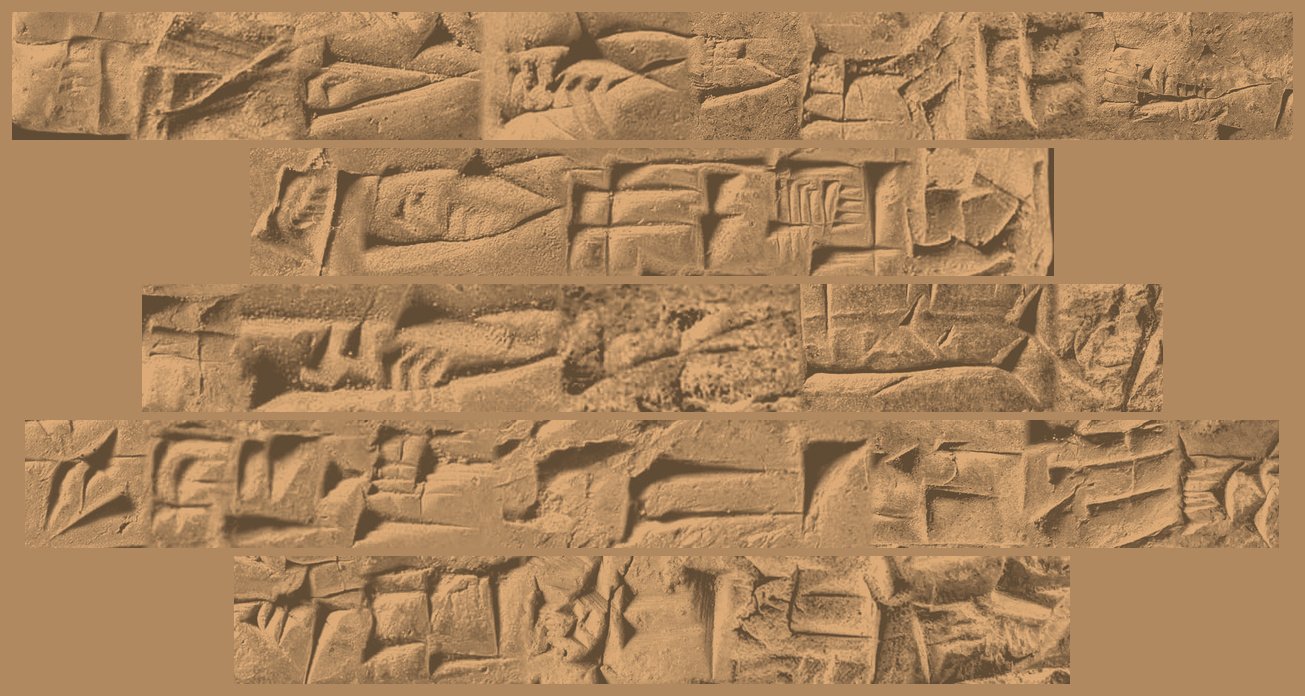}
  & \raggedright Autoregressive tablet-side transcription
  & \raggedright Encoder, decoder, tablet-side projector
  & \raggedright Patch embedding; Stages 1--2 \tabularnewline
\midrule
5 & \raggedright Real tablet sides
  & \vspace{0pt}\includegraphics[width=\linewidth,height=0.55in,keepaspectratio]{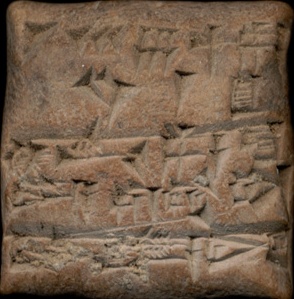}
  & \raggedright Autoregressive tablet-side transcription
  & \raggedright Encoder, decoder, tablet-side projector
  & \raggedright None for \(2\to5\), \(5\); otherwise patch embedding and Stages 1--2 \tabularnewline
\bottomrule
\end{tabular}
\end{table}

\paragraph{Image preparation and targets.}
Images are converted to RGB, resized while preserving aspect ratio, and centered on a \(512 \times 1024\) canvas. Maximum content bounds are \(256 \times 256\) pixels for sign crops, \(128 \times 1024\) for lines, and \(512 \times 512\) for tablet sides. The remaining canvas is filled with the image's median color, and pixel values are normalized using ImageNet channel means and standard deviations. A visual attention mask excludes padded regions.

\paragraph{Optimization.}
Epoch counts and batch sizes are listed in Table~\ref{tab:training-schedule}. Each phase uses a fresh AdamW optimizer with a constant learning rate of \(10^{-4}\), weight decay of \(0.01\), and gradient clipping at norm 1.0. We train in BF16 on an NVIDIA GeForce RTX 5070 Ti for seeds 13 and 42 and on an RTX 5090 for seed 123. We compute teacher-forced validation metrics after each image-training epoch and report the final Phase-5 checkpoints for curriculum comparisons.

\begin{table}[t]
\centering
\caption{Fixed settings for each phase when included in a curriculum.}
\label{tab:training-schedule}
\begin{tabular}{lrrc}
\hline
Phase & Epochs & Batch size & Relative Spatial Resolution \\
\hline
1: Sign recognition       & 10 & 12 & /32 \\
2: Text warm-up           &  1 & 32 & --- \\
3: Synthetic lines        & 10 & 32 & /16 \\
4: Synthetic tablet sides & 10 &  6 & /16 \\
5: Real tablet sides      & 25 &  9 & /16 \\
\hline
\end{tabular}
\end{table}

\FloatBarrier

\section{Evaluation}
\label{sec:evaluation}

We evaluate transcription by free generation and compare curricula using the final Phase-5 checkpoints (epoch 25). Spatial grounding is evaluated on a separate annotated holdout by measuring cross-attention alignment with ground-truth sign boxes during free generation, in addition to controlled occlusion.

\paragraph{Sign Error Rate}
After removing structural, uncertainty, and model-control tokens, we compute overall SER as the total minimum edit count divided by the total reference sign count, and mean tablet-side SER as the average of these ratios across tablet sides. 
\[
\mathrm{SER} =
\frac{\sum_{i=1}^{N}(S_i+D_i+I_i)}{\sum_{i=1}^{N}L_i},
\qquad
\mathrm{Mean\ side\ SER} =
\frac{1}{N}\sum_{i=1}^{N}\frac{S_i+D_i+I_i}{L_i},
\]
where \(N\) is the number of tablet sides, \(L_i\) is the reference sign count, and \(S_i,D_i,I_i\) are substitution, deletion, and insertion counts under a minimum-edit alignment.

\paragraph{Macro Sign F1}
We match predicted and reference sign sequences using a minimum-edit alignment, then compute F1 for each sign class and average across classes. This gives rare and frequent classes equal weight. A class that is never predicted receives an F1 of zero. 
\[
F_{1,c} = \frac{2TP_c}{2TP_c + FP_c + FN_c},
\qquad
\mathrm{Macro}\ F_1 = \frac{1}{|\mathcal{C}|}
\sum_{c\in\mathcal{C}} F_{1,c},
\]
where \(\mathcal{C}\) contains the sign classes present in the reference set, and \(TP_c\), \(FP_c\), and \(FN_c\) count correctly matched, incorrectly predicted, and missed instances of class \(c\), respectively.

\paragraph{Localization}
We measure localization using final-layer cross-attention averaged
over heads during autoregressive transcription. Predicted signs are
aligned with the reference using minimum-edit alignment. Peak in box
measures how often the model's strongest attention falls inside the
correct sign's box, while mass in box measures how much of its total
attention falls inside that box. Substitutions and deletions contribute zero.
\[
\mathrm{Peak\ in\ box} =
\frac{100}{N}\sum_{i\in\mathcal{C}}
\mathbf{1}\!\left[\operatorname*{arg\,max}_{p} A_i(p)\in B_i\right],
\qquad
\mathrm{Mass\ in\ box} =
\frac{100}{N}\sum_{i\in\mathcal{C}}\sum_p A_i(p)\,w_i(p),
\]
where \(N\) is the total number of eligible annotated reference signs,
including substitutions and deletions, and \(\mathcal{C}\) contains
the reference positions aligned with correctly recognized signs.
For \(i\in\mathcal{C}\), \(A_i(p)\) is normalized attention at image
pixel \(p\) for the corresponding generated sign, \(B_i\) is the
annotated sign box, and \(w_i(p)\) is the fraction of pixel \(p\)
covered by \(B_i\). Higher values indicate stronger alignment.

\paragraph{Controlled occlusion}
We mask either the target sign or an equal-size, nonoverlapping control region matched for foreground coverage, holding preceding generated tokens fixed. On the localization test set, we measure excess probability drop:
\[
\Delta_{\mathrm{excess}} =
\frac{100}{M}\sum_{i=1}^{M}
\left(q_i^{\mathrm{control}}-q_i^{\mathrm{target}}\right).
\]
Here, \(q_i\) is the reference-sign probability under each mask and \(M\) counts eligible pairs. Positive values indicate greater harm from target masking, measured in percentage points. Mask construction and exclusions are detailed in Appendix~\ref{app:occlusion-pairs}.

\FloatBarrier

\section{Results and Discussion}

\begin{table}[!htbp]
\centering
\caption{Test results for final Phase-5 epoch-25 checkpoints and fine-tuned baselines.
Values are percentages except excess probability drop (percentage points), reported as mean $\pm$ sample standard deviation
across seeds 13, 42, and 123, except the single-run Phase-5-only control
(seed 13) and PaddleOCR-VL (seed 42).
Training cost and structural diagnostics are reported in
Appendix~\ref{app:cost-diagnostics}.}
\label{tab:curriculum-results}
\small
\setlength{\tabcolsep}{4pt}
\resizebox{\textwidth}{!}{%
\begin{tabular}{lrrrrrr}
\toprule
\textbf{Model / phase sequence} &
\shortstack{\textbf{Pooled}\\\textbf{SER}} &
\textbf{Macro F1} &
\shortstack{\textbf{Mean side}\\\textbf{SER}} &
\shortstack{\textbf{Peak in}\\\textbf{box}} &
\shortstack{\textbf{Mass in}\\\textbf{box}} &
\shortstack{\textbf{Excess probability}\\\textbf{drop}} \\
\midrule
$1\to2\to3\to4\to5$
& $30.01 \pm 1.03$
& $54.58 \pm 0.83$
& $26.73 \pm 0.51$
& $70.99 \pm 2.09$
& $46.79 \pm 2.75$ & $61.73 \pm 1.34$ \\
$1\to2\to5$
& $31.37 \pm 0.65$
& $53.06 \pm 0.76$
& $27.12 \pm 0.13$
& $65.19 \pm 4.90$
& $42.44 \pm 3.61$ & $56.85 \pm 0.41$ \\
$2\to5$
& $29.40 \pm 0.66$
& $54.48 \pm 0.73$
& $\mathbf{25.68 \pm 0.48}$
& $51.27 \pm 15.40$
& $27.07 \pm 7.99$ & $61.58 \pm 1.44$ \\
$1\to2\to3\to5$
& $30.11 \pm 0.35$
& $53.88 \pm 0.33$
& $26.09 \pm 0.23$
& $66.93 \pm 4.45$
& $42.97 \pm 3.78$ & $60.95 \pm 2.81$ \\
$1\to2\to4\to5$
& $30.45 \pm 1.46$
& $52.99 \pm 1.39$
& $26.84 \pm 1.02$
& $65.23 \pm 8.21$
& $43.71 \pm 5.59$ & $61.54 \pm 2.48$ \\
$5$
& $105.20$
& $0.96$
& $108.06$
& $0.00$
& $0.21$ & $0.004$ \\
\midrule
eBL DETR (fine-tuned)
& $77.08 \pm 0.05$
& $10.26 \pm 0.03$
& $77.63 \pm 0.03$
& -- & -- & -- \\
PaddleOCR-VL (LoRA)
& $\mathbf{29.16}$
& $\mathbf{55.65}$
& $30.16$
& $\mathbf{83.96}$ & $\mathbf{53.73}$ & $66.35$ \\
\bottomrule
\end{tabular}%
}
\end{table}

\paragraph{Transcription}
When only looking at transcription performance, the results suggest that the specific curriculum matters much less than whether the model is given any pretraining before Phase-5. The five curriculum-trained models all end up in a narrow performance range, with SER around 30\% and macro F1 around 50\%. For the three seed-13 runs, Phase-5 validation losses and teacher-forced SERs also converge to nearly the same values by the end of training. This suggests that the earlier phases mainly change the model's initialization and optimization trajectory, while continued training on the shared Phase-5 objective pulls the models toward a similar final solution. In other words, once the model has learned how to extract useful information from the image, the common final objective appears to dominate the exact path used to get there. 

This is supported by the strong performance of the fine-tuned PaddleOCR-VL model, which achieved a similar pooled SER of 29.16\%. Its OCR pretraining likely provided transferable visual recognition and localization capabilities that support learning from real tablet-side data. However, all five pretrained TabletVLM curricula achieved lower mean side SER and comparable macro F1 while using approximately 3\% of its parameters. This demonstrates that a compact model can provide competitive transcription performance in this data-scarce domain without the computational cost of a much larger VLM.

In contrast, the Phase-5-only model was only pretrained on ImageNet. Even though the training and validation losses steadily decreased, its image-identity sensitivity remained essentially zero, which means it is not using the image tokens to determine the output. This strongly suggests that the model has learned to predict primarily from previously supplied target tokens only. The pretraining seems to serve primarily as a mechanism for preventing this language-only shortcut and providing a more favorable initialization before full tablet-side training begins.

All five pretrained curricula also substantially outperform the fine-tuned eBL pipeline, which achieves $77.08\pm0.05$\% pooled SER (Appendix~\ref{app:baseline-setup}). This was expected given the limitations of our sign-level annotations and is consistent with the performance of a similar detector-based pipeline we initially built and trained on a smaller dataset.

\paragraph{Localization}
The main difference between the earlier curriculum stages is in localization performance. The $1\to2\to3\to4\to5$ curriculum has the highest three-seed mean, achieving $70.99\%$ free-generation peak attention in box and $46.79\%$ attention mass in box. The $1\to2\to5$ curriculum achieves slightly lower localization scores of $65.19\%$ peak attention in box and $42.44\%$ attention mass in box. The $2\to5$ curriculum shows substantial localization variability, with free-generation peak-in-box ranging from 40.05\% to 68.83\%. The difference between these shorter curricula may reflect sign-level training as well as which encoder layers are trained in Phase 5. PaddleOCR-VL achieves higher peak-in-box (83.96\%) and mass-in-box (53.73\%).

The synthetic-stage ablations further highlight differences in localization stability. Replacing synthetic-line training in $1\to2\to3\to5$ with synthetic tablet-side training in $1\to2\to4\to5$ increases the free-generation peak-in-box standard deviation from 4.45 to 8.21 percentage points and mass-in-box standard deviation from 3.78 to 5.59 points. Adding the tablet-side stage after the line stage raises mean peak-in-box from 66.93\% to 70.99\% and lowers its standard deviation from 4.45 to 2.09 points. Across the three seeds, this pattern suggests that synthetic-line training may provide a useful bridge before learning from composed tablet layouts.

Additionally, masking the target sign region reduces TabletVLM's probability of the correct sign by 56.85--61.73 percentage points more than masking an equally sized control region across the five pretrained curricula. Previous predictions are held fixed across conditions. This indicates that predictions depend on the corresponding visual regions. PaddleOCR-VL shows the same pattern, with an excess drop of 66.35 points under its sign-probability scoring protocol (Appendix~\ref{app:occlusion-pairs}).

\paragraph{Performance Ceiling and Error Sources}
The similar pooled SER achieved by all five pretrained TabletVLM curricula and fine-tuned PaddleOCR-VL suggests that the remaining errors may partly reflect limitations of the shared dataset, such as variation in image quality and uncertainty in the transcriptions. For TabletVLM, the common Phase-5 data and training objective may matter more than the precise curriculum once sufficient visual grounding has been established.

\paragraph{Image resolution}
A controlled study is needed to determine whether higher input resolution improves transcription and localization on densely written tablet sides. In preliminary experiments, tablet-side training converged after we reduced the encoder stride from 32 to 16. Across six stride-16 runs, the fifth of tablet sides with the most reference signs per visible encoder feature cell had a pooled SER of 39.8--45.2\%, compared with 17.5--24.5\% across the middle three fifths. Such a study is challenging because paired tablet photographs and transcriptions are scarce, and fewer still have sign boxes for evaluating localization. The available photographs also vary in native resolution. Comparing input sizes on the same subset of sufficiently detailed photographs would isolate the effect of resizing more clearly, but would further reduce the number of sides available for evaluation.

\begin{figure}[!htbp]
\centering
\includegraphics[width=\textwidth]{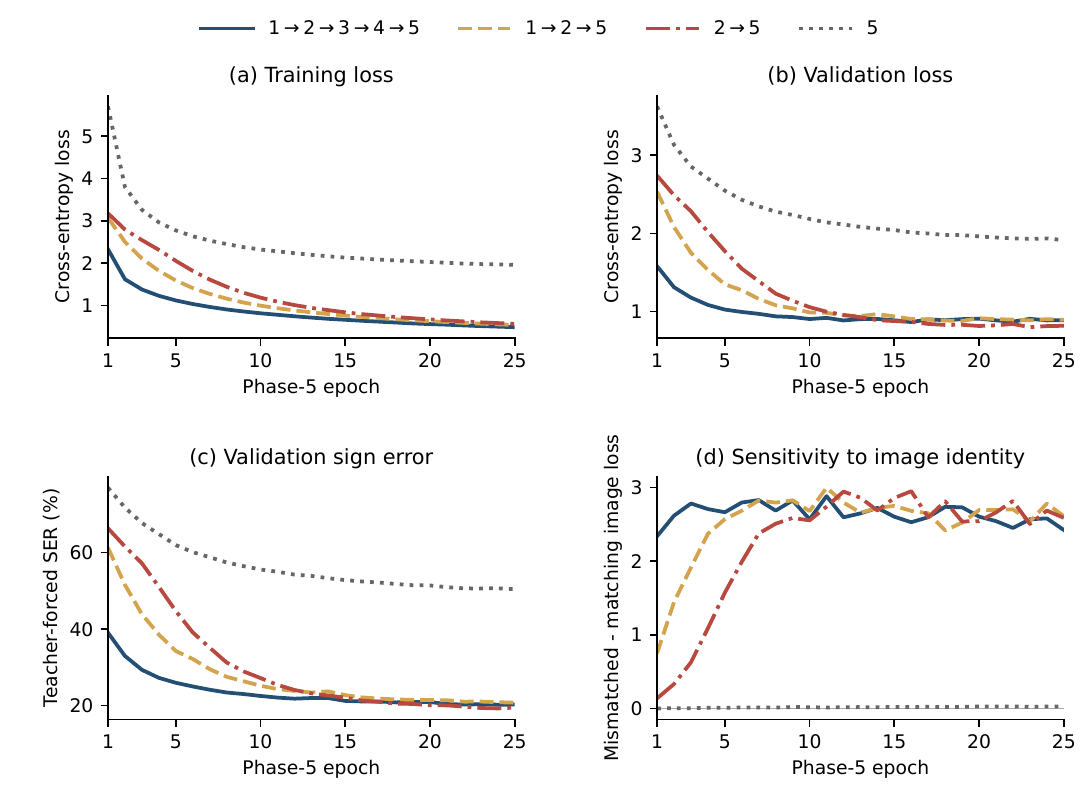}
\caption{Phase-5 training curves for the four seed-13 curricula: (a) training loss, (b) validation loss, (c) teacher-forced validation SER, and (d) the validation loss increase when images are mismatched with their transcriptions. A larger loss increase indicates greater sensitivity to image identity.}
\label{fig:training-curves}
\end{figure}

\begin{figure}[!htbp]
\centering
\includegraphics[width=\textwidth]{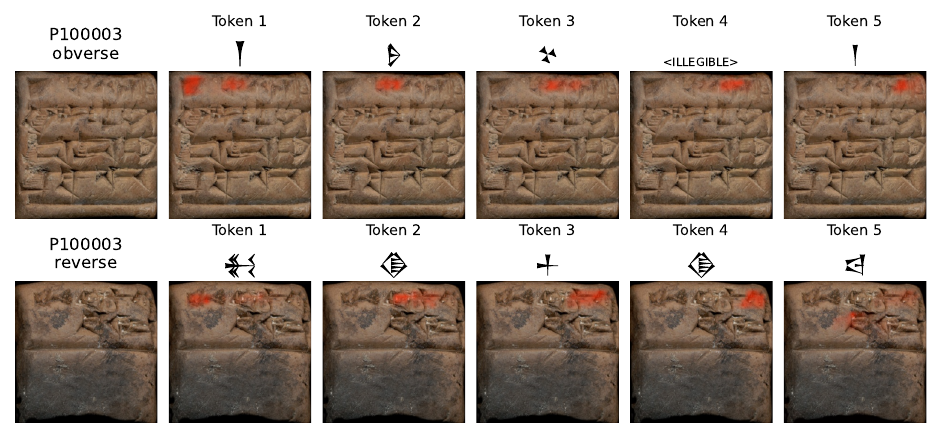}
\caption{Generated-token attention for the full-curriculum epoch-25 model on P100003. Each row shows the original side and the first five generated tokens, including structural markers. Red overlays show final-layer cross-attention averaged across heads, conditioned on generated prefixes and scaled to each token's peak (FP32 greedy decoding; opacity 0.65). These qualitative examples complement the free-generation localization evaluation.}
\label{fig:generated-attention}
\end{figure}

\FloatBarrier

\section{Conclusion}
We developed a single vision-language model for cuneiform tablet transcription and evaluated curricula spanning sign recognition, text modeling, and real tablet-side transcription. We highlight the full curriculum because it achieves the highest mean free-generation localization scores among TabletVLM curricula while remaining close to the lowest pooled SER (29.40\% for $2\to5$). Across three seeds, the full curriculum achieves $30.01\pm1.03$\% pooled SER, compared with $77.08\pm0.05$\% for the fine-tuned eBL DETR cuneiform baseline. Compared with fine-tuned PaddleOCR-VL, TabletVLM achieves similar pooled SER and a mean side SER 3.43 percentage points lower, using approximately 3\% of the parameters.

Our results demonstrate that full tablet transcription and spatial grounding can be learned within a single image-to-text model. This model substantially outperforms the evaluated multi-stage cuneiform OCR baseline, in which localization, recognition, and sequencing are optimized separately and errors can propagate between stages. These results support its use in the cuneiform domain, where data is scarce and imperfect. The curriculum experiments further suggest that intermediate supervision at simpler visual and sequential scales can shape capabilities that are difficult to learn from full tablet images alone. Future work should therefore investigate stronger mechanisms for spatial grounding and evaluate whether the same model architecture and curriculum principles transfer across periods, scripts, and other low-resource document transcription settings.

Overall, these findings demonstrate that single-model image-to-sequence transcription can outperform a multi-stage cuneiform OCR pipeline by jointly learning visual recognition, spatial grounding, and sequence prediction in a low-resource domain.

\clearpage
\section*{Ethics Statement}
The study uses cuneiform images and transcriptions associated with CDLI, eBL, and Sign Snippets. Their provenance and reuse conditions should be respected when distributing derived data or model artifacts. Autoregressively generated transcriptions can hallucinate and should be treated as aids to scholarly work rather than authoritative editions of cultural heritage material. 

\section*{Reproducibility Statement}
Section~\ref{sec:data-sources} and Appendix~\ref{app:dataset-construction} describe data sources, filtering, normalization, source-level splits, and synthetic rendering. Sections~\ref{sec:training-curricula} and~\ref{sec:evaluation} specify the model, phase schedule, checkpoint selection, metrics, and localization holdout. Appendix~\ref{app:model-reproducibility} records implementation and artifact identities. The reported comparisons use the fixed manifests and seed described in Section~\ref{sec:training-curricula}.

\section*{AI Use Statement}
We used OpenAI Codex to assist with manuscript structure and editing, LaTeX tables, dataset figure code, and discussion of the reported comparisons. It was also used to research existing techniques, related works, and brainstorm potential research ideas. We take responsibility for the final text, figures, code, and numerical claims.
\bibliography{iclr2027_conference}
\bibliographystyle{iclr2027_conference}

\appendix
\section{Dataset construction details}
\label{app:dataset-construction}

\subsection{Unicode normalization}
\label{app:unicode-normalization}
We construct the Unicode mapping by extending the \texttt{Cdli2cuneiform} workflow\footnote{\url{https://github.com/HarrisonLiruiHuang/Cdli2cuneiform}}, which maps transliterated sign names and values to Unicode cuneiform characters. Candidate mappings are aggregated from Akkademia, the Oracc Sign List, CDLI, CuneiML, Nuolenna, Text-Fabric, and Wikidata. Candidate outputs are normalized by removing spaces and retaining only valid cuneiform code points. For each source entry, the most frequent valid Unicode output is selected. Ties are resolved pseudorandomly from sorted candidates using seed 0. Reading labels are subsequently lowercased, and duplicate labels are reduced using the same seeded procedure.

This procedure provides a deterministic mapping for model training but does not constitute a philological resolution of disagreements between sign lists. The saved mapping artifact is therefore treated as the canonical mapping used throughout the experiments. The tokenizer adds \texttt{<LINE>}, \texttt{<COLUMN>}, \texttt{<ILLEGIBLE>}, and \texttt{<GAP>} as structural tokens and \texttt{<PAD>}, \texttt{<BOS>}, \texttt{<EOS>}, and \texttt{<UNK>} as model-control tokens. Validation and transcription test sets contain 402 and 403 distinct sign characters, respectively, with no sign classes outside the training inventory.

\subsection{Sign Snippets filtering}
\label{app:sign-snippets}
We retain Sign Snippets examples assigned to the Ur III period and use Electronic Babylonian Literature (eBL) resources to recover their source images and catalog identifiers. After lowercasing and trimming \texttt{signName} and \texttt{value}, we remove bracketed sign names and names equal to \texttt{unknown}, \texttt{struct}, or \texttt{x}. We also remove rows whose two nonempty labels differ as strings and rows whose sign name cannot be mapped to Unicode. String inequality is a filtering heuristic, not evidence that the annotation is incorrect: different readings can map to the same sign.

Mappings between sign images and Unicode labels were visually inspected using a custom Streamlit interface. eBL identifiers were subsequently mapped to CDLI P-numbers and linked to their corresponding source tablets. These source identities are retained throughout dataset generation to prevent sign crops originating from held-out tablets from entering training examples.

Reapplying these filters to the saved full Sign Snippets metadata export begins with 158,946 records. Filtering removes 4,681 bracketed sign names, 755 \texttt{unknown}/\texttt{struct}/\texttt{x} entries, and 53,095 unequal name--value pairs, leaving 100,415 metadata records before Ur III and final source-eligibility filtering. Of the unequal pairs, 50,050 map to the same Unicode under our mapping. The final study manifest contains 4,964 retained Ur III sign crops across the training and validation splits.

The coverage comparison is restricted to tablets represented by retained sign rows in the train/validation manifest; it does not estimate coverage for all eBL tablets. The denominator counts visible sign positions in eBL text lines, excluding broken-away signs, gaps of unknown length, and structural markers. Path-aligned coverage counts distinct positions with a box at the same eBL transcription path; stale paths may lower this estimate. Boxes without an identifiable side or a visible-sign denominator are omitted from the corresponding ratios.

\subsection{CDLI ATF parsing}
CDLI catalog records provide ATF transcriptions and metadata, while composite photographs are retrieved by P-number. We restrict the study dataset to Ur III tablets and match photographs to transcription records using these identities. A paired example requires an available photograph, a non-empty side transcription, and period metadata.

We parse ATF side directives for the obverse, reverse, left, right, top, and bottom surfaces and normalize recognized spelling and punctuation variants in surface names. Document headers, comments, and other non-transcription metadata are removed. Editorial line numbers, including primed, ranged, and subdivided labels, are stripped, and indented continuation text is joined to its preceding line.

Transliterations are segmented at hyphens and determinative braces while retaining the enclosed determinative text. Editorial punctuation, inline status annotations, and allograph suffixes are removed before readings are lowercased. Tokenization preserves parenthesized numeric expressions and pipe-delimited compound sign names while separating compound operators outside these groups.

Parsed rows are separated by \texttt{<LINE>}, while column directives are represented using \texttt{<COLUMN>}. Bracketed spans and isolated \texttt{x} readings are mapped to \texttt{<ILLEGIBLE>}. ATF status markers explicitly describing illegibility are likewise represented by \texttt{<ILLEGIBLE>}, whereas broken, damaged, effaced, missing, or trace-only content is represented by \texttt{<GAP>}. Normalized sign readings are then converted to Unicode, with unmapped labels represented by \texttt{<ILLEGIBLE>}. We retain tablet-side targets containing between 1 and 100 Unicode cuneiform signs, excluding structural markers from the sign count.

\subsection{CDLI tablet-side extraction}
Although the transcription parser recognizes multiple tablet surfaces, automatic image extraction is restricted to obverse and reverse sides. Composite photographs are first converted to grayscale for region detection. To identify photographs with the expected black background, we inspect a square patch at each corner with side length equal to the larger of 16 pixels or 8\% of the shorter image dimension. An image is retained only when at least 95\% of pixels in every corner have grayscale intensity at most 50 on a 0--255 scale.

Pixels with intensity greater than 50 are treated as foreground, and 8-connected components are extracted using OpenCV. Components are retained when their foreground area occupies 0.8--70\% of the complete image and their bounding-box width-to-height ratio lies between 0.10 and 12.0. The two largest eligible components are selected by foreground area. Assuming the standard CDLI composite layout, the upper component is assigned to the obverse and the lower component to the reverse; photographs with fewer than two eligible components are discarded.

Each selected component is cropped from the original color photograph and stored as a JPEG at quality 95. The foreground mask is used only to determine the crop boundary, so black pixels falling within the bounding rectangle are retained. Only successfully extracted sides with corresponding normalized transcription records are included in the manifest.

\subsection{Synthetic rendering}
\label{app:synthetic-rendering}
\paragraph{Synthetic lines.}
Within each split and period, we divide CDLI transcriptions at line and column boundaries and retain sequences of 5--10 signs for which eligible sign crops are available. We sample a real crop for each sign and compose the crops in transcription order into a labeled synthetic line.

\paragraph{Synthetic tablet sides.}
Within each split and period, we interleave lines from different source tablets and group four to eight lines into a synthetic tablet side. These compositions do not reconstruct the discourse or layout of a physical tablet; their target sequences are joined with \texttt{<LINE>} tokens.

\begin{figure}[t]
\centering
\includegraphics[width=0.8\linewidth]{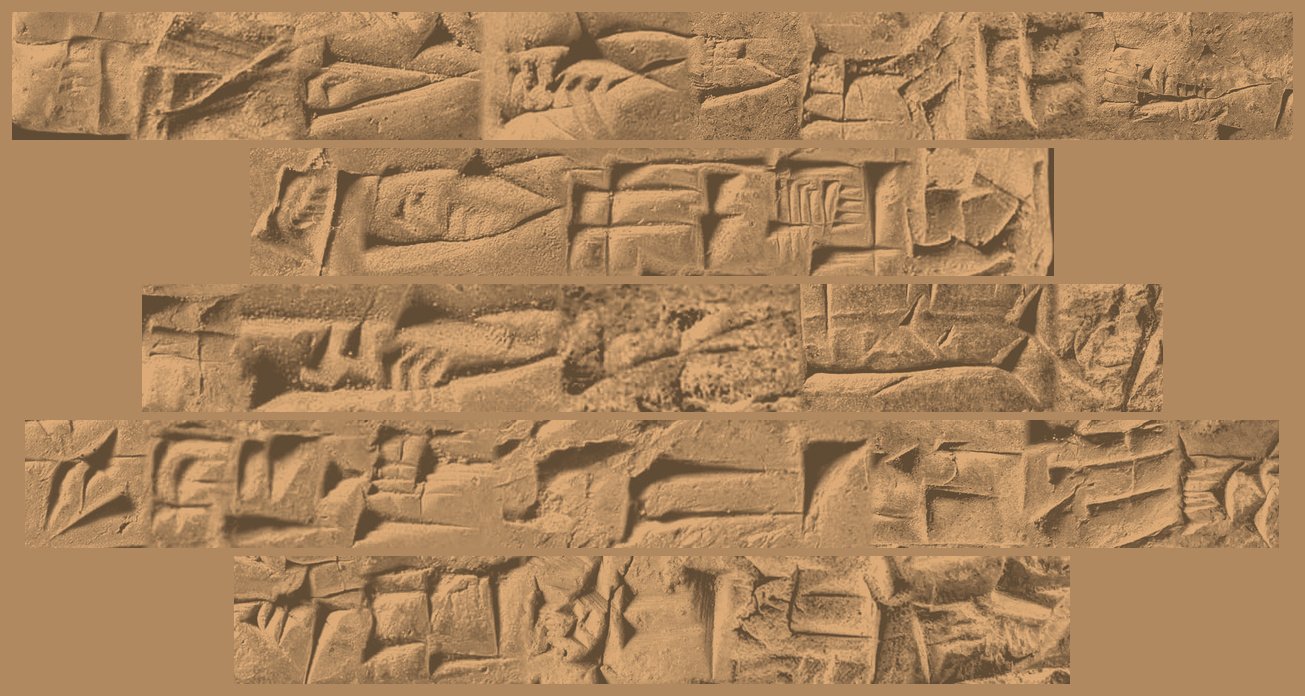}
\caption{Example synthetic tablet side assembled from five synthetic lines. The sign crops come from real tablets, but the composition does not represent a physical tablet.}
\label{fig:synthetic-tablet-side}
\end{figure}

When duplicate tablet-side transcription records exist, we retain the record containing the largest number of Unicode signs, with stable ordering used to resolve ties. Lines containing remaining control tokens or signs without an eligible crop in the same split and period are discarded. For every sign in a retained sequence, a crop is sampled from the same split and period using seed 0 and sorted candidate lists. Crops may be reused between examples, and each eligible source line generates one synthetic image.

Each crop is converted to grayscale and intensity-normalized using its 5th and 95th percentiles, clipping the result to $[0,1]$. Each normalized pixel intensity $u$ deterministically scales the base clay color RGB $(176,137,96)$ by $0.55+0.70u$; constant-intensity crops use the base clay color directly. Crops are resized to a height of 128 pixels using Lanczos interpolation while preserving aspect ratio. They are placed horizontally in transcription order with up to 8 pixels of overlap, using linear blending at crop boundaries. Synthetic lines are stored as RGB JPEGs at quality 95 with chroma subsampling disabled.

Synthetic lines are shuffled within each source tablet, and source-tablet groups are interleaved using seed 0. The ordered lines are partitioned into groups of four to eight, subject to a nominal maximum of 100 signs. Because individual lines contain at most 10 signs, an eight-line composition contains at most 80 signs. Lines are stacked vertically without additional resizing, horizontally centered, and separated by 8-pixel gaps with 12 pixels of outer padding on the same RGB $(176,137,96)$ background. Synthetic sides are stored as JPEGs at quality 95 with chroma subsampling disabled.

\subsection{Source assignment and dataset accounting}
\label{app:source-accounting}
The fixed transcription test set is inherited from a reference manifest and is not resampled. Every test P-number is excluded from training and validation, including other tablet sides, sign crops, and transcription donors from that tablet. After removing test and localization-holdout identities, 6,470 of 64,699 candidate source identities are assigned to validation and 58,229 to training. Only 4,398 validation identities occur in retained examples because final synthetic inclusion also requires eligible crops and transcriptions; 10\% describes source assignment, not the final row proportion.

Validation assignment follows SHA-256 ranking but skips a candidate if removing it would eliminate the last training representative of a period or sign within a real-image level and period. It therefore preserves training coverage rather than taking an unconstrained hash-ranked split.

Each synthetic line records the P-number supplying its transcription and every sign crop; each synthetic side records the union of its constituent lines' identities. An example is retained only if all donors belong to its split and period. Decoder text warm-up uses only training-split transcriptions. The final manifest has zero pairwise source-identity overlap across training, validation, and transcription test splits.

Across all components, the training, validation, and test splits contain 58,229, 4,398, and 2,485 unique recorded source P-numbers, respectively. These counts include transcription and sign-image donors and are not counts of photographed tablet sides. Considering only real CDLI tablet-side examples, the corresponding counts are 20,117, 2,198, and 2,485. The rebuilt manifest preserves all 4,964 eligible real sign crops and all 43,176 real tablet sides from the training/validation pool. It contains 145,418 synthetic lines and 18,178 synthetic sides across those splits; the frozen transcription test set contains 4,805 sides.

In the full curriculum, synthetic training pools are capped at 10,000 lines and 10,000 synthetic sides, while all 4,520 training sign crops and 38,920 real training tablet sides are used. The image-training pool therefore contains 63,440 distinct examples before repeated epochs. Text-only decoder warm-up uses the uncapped training transcription pool. Manifest totals are not the number of examples observed in every curriculum stage.

Before phase caps, the training sign crops cover 167 sign classes and each synthetic training pool covers 166; real training sides cover 470 of the combined 471 classes. The intermediate image sources therefore cover only part of the output vocabulary.

\subsection{Label coverage and uncertainty}
The training manifest contains 2,855,151 cuneiform sign-token occurrences, compared with 173,947 in validation and 131,277 in the transcription test set. These totals include repeated textual content appearing in synthetic lines and synthetic tablet sides.

Table~\ref{tab:label_uncertainty} summarizes uncertainty markers for the real CDLI tablet-side subset. Percentages are computed over cuneiform signs together with the four structural-marker classes, excluding model-control tokens. Because \texttt{<ILLEGIBLE>} represents both damaged or unknown content and readings that could not be mapped to Unicode, its frequency should not be interpreted solely as a mapping-failure rate.

\begin{table}[t]
\centering
\caption{Label uncertainty in real CDLI tablet-side transcriptions.}
\label{tab:label_uncertainty}
\begin{tabular}{lrrrrr}
\toprule
\textbf{Split} &
\textbf{Signs} &
\textbf{Illegible} &
\textbf{Share} &
\textbf{Gap} &
\textbf{Share} \\
\midrule
Train
    & 1,084,771 & 76,007 & 5.68\% & 2,244 & 0.17\% \\
Validation
    & 118,849 & 7,933 & 5.43\% & 234 & 0.16\% \\
Test
    & 131,277 & 9,142 & 5.65\% & 264 & 0.16\% \\
\bottomrule
\end{tabular}
\end{table}

\subsection{Localization holdout details}
\label{app:localization-holdout}
The 67-tablet localization holdout was reserved from 133 tablets containing labeled sign coordinates. All holdout tablets have resolved CDLI identities, including documented joins, stored in an exclusion ledger applied before synthetic generation and train--validation assignment. The exclusions cover real tablet sides, sign crops, synthetic components, and transcription donors.

The complete frozen localization population contains 283 artifacts, of which 279 were processed and four remain unresolved. The 67-tablet subset contains 178 side records. Ninety-nine sides have no eligible localization targets, leaving 79 sides with 882 eligible annotated positions; three out-of-vocabulary positions are excluded for evaluation of the final full-curriculum model, leaving 879 scored positions. No retained localization side exceeds the model context limit.

Localization uses separate eBL side crops. Verified sign anchors identify a unique connected photograph region; explicit surface annotations are used when available. The region is expanded to retain complete sign boxes. Stale transcription paths, changed readings, conflicting or ambiguous assignments, unresolved crops, and sides with transcription issues are excluded from scoring.

\paragraph{Attention geometry}
For both localization protocols, final-layer head-averaged attention is bilinearly upsampled to the input canvas, cropped to the resized image content, and renormalized to sum to one. Peak-in-box uses the maximum pixel's center; mass-in-box weights boundary pixels by fractional box overlap. Both metrics average equally over the 879 scored positions. Uniform attention over image content yields a mean mass-in-box of 3.58\%.

\paragraph{Free-generation protocol}
The main results use greedy FP32 generation from \texttt{<BOS>}, conditioned only on the model's own preceding predictions, stopping at \texttt{<EOS>} or the 512-token limit. We replay the generated prefixes through the causal decoder in FP32 to extract attention at the steps predicting generated signs; this replay contains no reference tokens. Structural tokens are retained in generation and replay. Reference and generated sign sequences are then aligned by unit-cost Levenshtein alignment, with diagonal, insertion, and deletion traceback priority. Correctly matched signs supply attention maps; substitutions and deletions score zero over the full reference denominator, while insertions have no reference box and receive no separate spatial penalty. Cached generation and full-prefix replay can differ by numerical rounding.

\paragraph{Reference-conditioned protocol}
As a supplementary diagnostic, attention is evaluated in FP32 using \texttt{<BOS>} and the canonical reference sequence, preserving out-of-vocabulary positions as \texttt{<UNK>}. This measures spatial alignment given the reference history rather than end-to-end recognition and localization. Table~\ref{tab:reference-localization} reports these scores separately from the free-generation results in Table~\ref{tab:curriculum-results}.

\begin{table}[htbp]
\centering
\caption{Reference-conditioned localization on the same 879 annotated test positions. Values are percentages, reported as mean $\pm$ sample SD across seeds 13, 42, and 123, except the seed-13 Phase-5-only control.}
\label{tab:reference-localization}
\begin{tabular}{lrr}
\toprule
Phase sequence & Peak in box & Mass in box \\
\midrule
$1\to2\to3\to4\to5$ & $80.51\pm4.94$ & $52.06\pm4.46$ \\
$1\to2\to5$ & $76.53\pm4.05$ & $48.71\pm3.64$ \\
$2\to5$ & $57.19\pm17.08$ & $30.09\pm9.04$ \\
$1\to2\to3\to5$ & $75.50\pm2.33$ & $47.89\pm3.29$ \\
$1\to2\to4\to5$ & $76.72\pm7.97$ & $50.33\pm5.87$ \\
$5$ & 1.48 & 3.30 \\
\bottomrule
\end{tabular}
\end{table}

Forty-four localization-holdout P-numbers occur in the transcription test set, corresponding to 83 transcription-test sides. These sides remain in the frozen transcription benchmark, and none of their known source identities occur in the final training or validation manifest. Exclusions apply to datasets rebuilt after the localization reservation; earlier-model localization measurements are diagnostic unless their training provenance establishes the same isolation. Held-out claims are associated with the exact training manifest of each checkpoint.

Source-level isolation controls overlap through known catalog identities and documented joins, but does not guarantee independence from unidentified catalog aliases or external image-encoder pretraining. Tablet-level separation also does not imply sequence-level uniqueness: recurring expressions can appear on distinct tablets, and identical textual subsequences are not removed across splits.

\paragraph{Controlled-occlusion population and masks}
\label{app:occlusion-pairs}
The reported excess probability drops use the held-out localization test set and fixed clean-image generated prefixes. Each run begins with 879 eligible annotated reference positions. Aligned substitutions remain eligible; deletions without a prediction step are excluded. Table~\ref{tab:occlusion-pairs} reports the retained pairs for each seed. All exclusions in these runs are deletions; no pairs were excluded for lack of a suitable control. Pair populations consequently differ across models and seeds. The results measure visual dependence under fixed textual context, rather than transcription after regenerating predictions on masked images.

Target boxes are projected onto resized image content using floor for left/top and ceiling for right/bottom boundaries. Masked pixels are replaced with the original image's median RGB color, with resizing, padding, and normalization held fixed. The control is equal in size, disjoint from the target, and differs in foreground fraction by at most five percentage points; foreground is grayscale intensity greater than 50. The farthest eligible control is selected by squared top-left distance with raster-order tie breaking, independently of model predictions. Controls may cover other signs.

Both models use the same masks on the resized images and score in FP32. TabletVLM scores one token per sign, suppressing \texttt{<PAD>} and \texttt{<BOS>} before softmax. PaddleOCR-VL scores the product of four conditional byte-token probabilities for the reference sign after the fixed generated prefix, using the full vocabulary at each step. The PaddleOCR-VL run retains 850 of 879 signs and yields an excess probability drop of 66.35 percentage points. The scoring conventions and retained reference positions differ across models, so the comparison is descriptive.

\begin{table}[htbp]
\centering
\caption{Eligible occlusion pairs retained from 879 annotated test positions under generated prefixes. Dashes indicate unevaluated seed settings.}
\label{tab:occlusion-pairs}
\begin{tabular}{lrrr}
\toprule
Model / phase sequence & Seed 13 & Seed 42 & Seed 123 \\
\midrule
$1\to2\to3\to4\to5$ & 841 & 846 & 840 \\
$1\to2\to5$ & 840 & 838 & 833 \\
$2\to5$ & 826 & 831 & 818 \\
$1\to2\to3\to5$ & 833 & 836 & 836 \\
$1\to2\to4\to5$ & 818 & 817 & 849 \\
$5$ & 593 & -- & -- \\
PaddleOCR-VL (LoRA) & -- & 850 & -- \\
\bottomrule
\end{tabular}
\end{table}

\section{Model implementation and reproducibility}
\label{app:model-reproducibility}

\subsection{Disabling Stage 4 spatial downsampling}
\label{app:swin-downsampling}
Standard SwinV2 patch merging concatenates the feature vectors from each non-overlapping $2\times2$ neighborhood, then applies a learned linear reduction and normalization. At the entrance to Stage 4, this maps an $H\times W\times384$ feature map to $(H/2)\times(W/2)\times768$.

During Phases 3-5, we replace neighborhood concatenation with four channelwise copies of the feature vector at each spatial position. For a vector $x_{h,w}\in\mathbb{R}^{384}$, the modified operation is
\[
y_{h,w}=\operatorname{LayerNorm}\!\left(W_{\mathrm{merge}}
[x_{h,w};x_{h,w};x_{h,w};x_{h,w}]\right),
\]
where $W_{\mathrm{merge}}$ is the pretrained linear reduction from 1,536 to 768 channels. The reduction and normalization retain their pretrained parameters at initialization. This preserves the $H\times W$ spatial grid while producing the 768 channels expected by the Stage 4 transformer blocks. For a $512\times1024$ input, the final grid is therefore $32\times64$ rather than $16\times32$ (Table~\ref{tab:swin-dimensions}). Phase 1 retains standard patch merging.

\subsection{Projection, position embeddings, and training}
The feature adapters and projection layers use LayerNorm, a linear map, and dropout. Each visual pathway has learned row and column embeddings initialized to zero; the decoder has learned sequence-position embeddings. Phase 1 classifies mean-pooled encoder features with a separate head. Valid visual tokens are packed before decoder attention. Last-epoch weights carry between phases, while validation-best checkpoints are retained separately.

\subsection{Checkpoints and decoding}
All reported curriculum comparisons use the checkpoint from epoch 25 of Phase 5. Transcription evaluation uses PyTorch \texttt{2.11.0+cu128} and batch-one greedy decoding in BF16, suppressing \texttt{<PAD>} and \texttt{<BOS>} after initialization; localization and controlled occlusion use FP32. Each experiment retains its dataset manifest, vocabulary, training history, and evaluation summaries, including checkpoint hashes and decoding settings.

\subsection{Seed and hardware provenance}
Seeds 13 and 42 were run on an RTX 5070 Ti; seed 123 was run on an RTX 5090 with 32\,GB memory. Seed 13 for $1\to2\to3\to4\to5$, $1\to2\to5$, and $2\to5$ uses the earlier implementation; the other seed-13 ablations and seed-42 runs use the updated local implementation. Seed-123 training uses revision \texttt{72d2ef75cff24a51b1d635e9a8950ef834b081a0} plus the environment-controlled seed patch, with PyTorch \texttt{2.11.0+cu128}, timm \texttt{1.0.27}, and Transformers \texttt{5.6.2}. Seed-123 localization and occlusion use the same evaluator revision without uncommitted changes to the evaluator or runtime. The three-seed summaries therefore span implementation revisions and hardware, with matching dataset and evaluation identities. A separate seed-13 RTX 5090 replication is excluded from the three-seed aggregates.

\subsection{Data versions and provenance}
The reported curricula share an identical dataset manifest and evaluation tokenizer. SHA-256 fingerprints of the manifest and tokenizer, including normalization settings, are recorded with the evaluation results.
The Sign Snippets dataset has a December 2025 Zenodo archive
(\url{https://doi.org/10.5281/zenodo.17949595}), distinct from the 2026
article~\citep{lewenstein2026cuneiform}; its image crops are distributed under
CC BY-NC 4.0. Localization uses a fixed eBL annotation export dated 11 September 2026, cached transcriptions and photographs, and recorded holdout membership and source exclusions.

\section{Baseline training and evaluation}
\label{app:baseline-setup}
All baselines use the same 4,805 held-out tablet-side images and shared transcription scorer. Checkpoints and inference settings are selected using validation data. Fine-tuned eBL results are mean $\pm$ sample standard deviation across seeds 13, 42, and 123; PaddleOCR-VL uses seed 42 due to its computational cost.

\subsection{eBL detector}
We fine-tune the released eBL detector of \citet{che2026automated} on fully boxed synthetic canvases assembled from training sign crops and font-rendered glyphs. Training uses 4,096 canvases per epoch for 10 epochs with AdamW, learning rate $2\times10^{-5}$, weight decay $10^{-4}$, gradient clipping 0.1, and effective batch size 16. At seed 42, validation pooled SER favors decoder-and-head fine-tuning over full-model fine-tuning; we repeat the selected configuration at seeds 13 and 123. Each run independently selects its checkpoint and confidence threshold on validation, yielding epoch 1 and threshold 0.30 for all three seeds. Inference uses aspect-preserving resizing to an 800-pixel short edge with a 1,333-pixel long-edge cap, up to 100 detections, and adapted eBL-style DBSCAN reading order. The detector receives box-supervised synthetic canvases, whereas TabletVLM also learns from real-side transcription pairs.

\subsection{PaddleOCR-VL}
We fine-tune PaddleOCR-VL-1.6 using LoRA on the 38,920 real training sides, with images fitted to a 512-pixel longest side. LoRA uses rank 16 and scaling parameter $\alpha=32$ on query, key, value, and output projections. Training uses BF16 precision, a learning rate of $2\times10^{-4}$ decaying to $2\times10^{-5}$ with a cosine schedule, batch size 2, gradient accumulation 2, and five epochs. The prompt is \texttt{OCR:}, and targets are the manifest transcriptions. Pooled SER on a fixed seeded 500-side validation subset is evaluated after each epoch and selects epoch 4. Test decoding is greedy, one side at a time, with a 1,024-token generation limit. Predicted cuneiform signs outside the training vocabulary are retained as errors; no such signs were generated on the test set.

\paragraph{Localization readout}
PaddleOCR-VL localization is evaluated in FP32 on the same 79 test sides and 879 annotated positions as TabletVLM, using the shared geometry and alignment scorer. Attention is averaged over heads and the queries predicting each sign's constituent byte tokens, retaining image-patch keys only. We compare all 18 layers with and without removal of each layer's attention sink: the image token receiving the greatest aggregate attention from text queries. Sink removal excludes that token and renormalizes the remaining attention; it does not alter generation. Peak-in-box on the annotated localization training split (97 sides, 1,302 positions) selects layer 13 with sink removal. This fixed readout yields 83.96\% free-generation peak-in-box and 53.73\% mass-in-box on test, versus 0.00\% and 0.60\% using final-layer mean-head attention. Under reference conditioning, the selected readout yields 92.04\% peak-in-box and 58.90\% mass-in-box. Training boxes select the readout only; test boxes are used for scoring. TabletVLM uses final-layer mean-head attention without layer selection or sink removal.

\section{Training cost and supplementary diagnostics}
\label{app:cost-diagnostics}

Table~\ref{tab:cost-diagnostics} reports saved Phase-5 diagnostics for seed 13. GPU-hours sum
completed training calls through epoch 25 across all included phases on one
RTX 5070 Ti. They exclude separate validation, evaluation, checkpoint writing,
preparation, and external pretraining, and are distinct from total run wall time.

Structural micro-F1 pools true positives, false positives, and false negatives
for the four structural markers after unit-cost alignment retaining signs and
markers; matches require identical marker types with no positional tolerance.
Generation-limit counts identify sides reaching 512 tokens without
\texttt{<EOS>}. These diagnostics capture structure and termination differences
excluded from sign-only metrics.

\begin{table}[htbp]
\centering
\caption{Seed-13 training cost and free-generation diagnostics on the 4,805 test sides.
Structural micro-F1 is a percentage; the final column counts sides.}
\label{tab:cost-diagnostics}
\begin{tabular}{lrrr}
\toprule
Curriculum & GPU-hours & Structural micro-F1 & Generation limit \\
\midrule
\(1\to2\to3\to4\to5\) & 7.765 & 55.43 & 65 \\
\(1\to2\to5\) & 6.766 & 64.42 & 34 \\
\(2\to5\) & 8.618 & 75.26 & 15 \\
\bottomrule
\end{tabular}
\end{table}

\end{document}

%% file: math_commands.tex
\usepackage{amsmath,amsfonts,bm}

\def\eqref#1{equation~\ref{#1}}

\def\1{\bm{1}}

\DeclareMathAlphabet{\mathsfit}{\encodingdefault}{\sfdefault}{m}{sl}
\SetMathAlphabet{\mathsfit}{bold}{\encodingdefault}{\sfdefault}{bx}{n}

